\documentclass{article}
\usepackage{ijcai22}

\usepackage{times}
\usepackage{soul}
\usepackage{url}
\usepackage{cite}
\usepackage[hidelinks]{hyperref}
\usepackage[utf8]{inputenc}
\usepackage[small]{caption}
\usepackage{graphicx}
\usepackage{float}
\graphicspath{{./images/}}
\usepackage{subfigure}
\usepackage{algorithm}
\usepackage{algorithmic}
\usepackage{amsmath}
\usepackage{booktabs}
\usepackage{indentfirst}
\usepackage{appendix}
\usepackage{xcolor}
\usepackage{natbib}
\newcommand{\figref}[1]{Figure~\ref{#1}}
\newcommand{\tabref}[1]{Table~\ref{#1}}

\newcommand{\AlgRef}[1]{Algorithm~\ref{#1}}
\newcommand{\equref}[1]{Equ.~(\ref{#1})}

\title{Learning Symbolic Operators: A Neurosymbolic Solution for Autonomous Disassembly of Electric Vehicle Battery}

\author{
Yidong Du$^{1,2}$\footnotemark[1]\and
Wenshuo Wang$^{1,2}$\footnotemark[1]\and
Zhigang Wang$^2$\footnotemark[2]\and
Hua Yang$^2$\and\\
Haitao Wang$^2$\and
Yinghao Cai$^1$\And
Ming Chen$^3$\footnotemark[2]\\
\affiliations
$^1$Institute of Automation, Chinese Academy of Sciences, China \\
$^2$Intel Labs China\\
$^3$School of Mechanical Engineering, Shanghai Jiao Tong University, Shanghai, China\\
\emails
jinglingniu123@outlook.com,
wenshuo\_wang@yeah.net,
\{zhi.gang.wang,hua.yang,hai.tao.wang\}@intel.com,
yinghao.cai@ia.ac.cn,
mingchen@sjtu.edu.cn
}

\begin{document}

\maketitle

\begin{abstract}
The booming of electric vehicles demands efficient battery disassembly for recycling to be environment-friendly. Currently, battery disassembly is still primarily done by humans, probably assisted by robots, due to the unstructured environment and high uncertainties. It is highly desirable to design autonomous solutions to improve work efficiency and lower human risks in high voltage and toxic environments. This paper proposes a novel neurosymbolic method, which augments the traditional Variational Autoencoder (VAE) model to learn symbolic operators based on raw sensory inputs and their relationships. The symbolic operators include a probabilistic state symbol grounding model and a state transition matrix for predicting states after each execution to enable autonomous task and motion plannings. At last the method's feasibility is verified through test results.
\end{abstract}

\section{Introduction}
\noindent With the booming of large-capacity lithium batteries, more automakers have adopted batteries to power electric cars~\cite{DIXON2020100059}. According to data from China Automotive Technology and Research Center (CATARC), in $2025$, the amount of expired electric car batteries will reach $780$k tons (about $116$GWh). End-of-life batteries pose hazards to the environment if not properly recycled. Leading-edge battery disassembly techniques rely on human-assisted machines, which not only are of low efficiency but also expose workers to haphazard working conditions. Having robots to disassemble batteries automatically is urgently needed to exempt humans from toxic working environments, improve cost efficiency, and handle heavy workloads. As of now, the major difficulties of using robots to autonomously disassemble batteries lie in the unstructured working environment and a multitude of uncertainties. Automating the process of disassembly, recycling and sorting battery components cannot be done by pre-programming against a homogeneous batch since end-of-life batteries are often of different models and shapes~\cite{harper2019recycling}. For automatic disassembly, several methods used visual servoing robots to carry out the task~\cite{li2021hierarchical,robotics10020082, met11030387}. However, these solutions mandate good enough sensory inputs for robots to detect the bolts, and the disassembly task planning is often ignored.

\begin{figure}
\includegraphics[height = 4.5cm,width=\columnwidth]{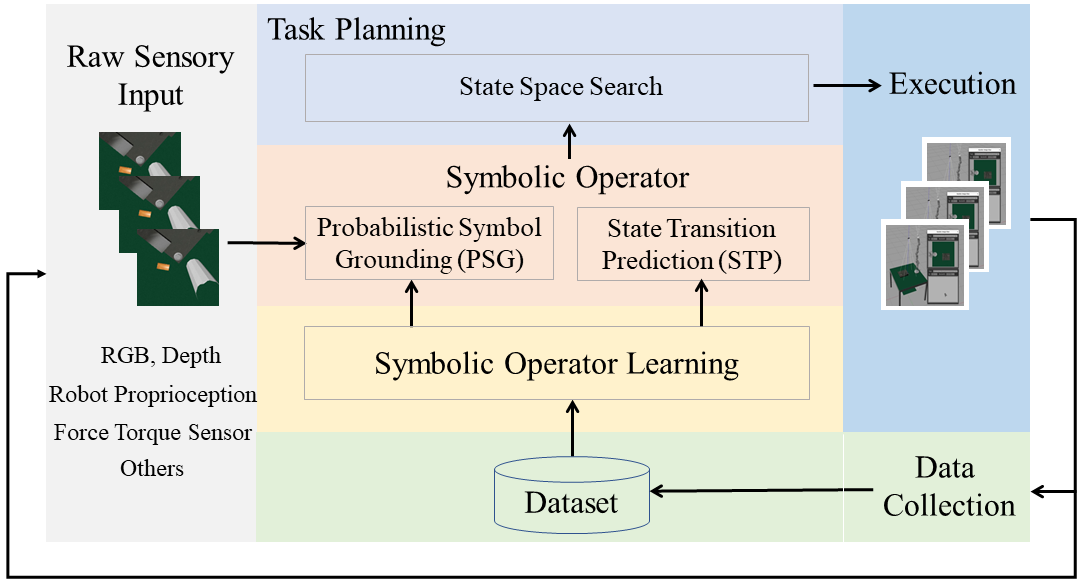}
\caption{NeuroSymbolic based Autonomous Disassembly System}\label{fig:arch}
\end{figure}

\figref{fig:arch} shows the proposed neurosymbolic based system framework. The proposed method trains symbolic operators, including a Probabilistic Symbol Grounding (PSG) model and a State Transition Prediction (STP) model, to enable autonomous task and motion planning (TAMP). Symbolic operators define a lossy abstraction of the transition model between the states. TAMP relies on these domain-specific symbolic operators to predict state transition under different actions and produce action sequences to reach the goal state. The system is able to plan online based on current environmental variations for autonomous execution. It can be easily extended with new capabilities for more complex tasks. This paper proposes a neurosymbolic method to disassemble batteries autonomously, subject to environment dynamics and uncertainties, based on symbolic operators learned from demonstrations.  Our main contributions are as follows: 
\begin{enumerate}
    \item Augment the traditional Variational Autoencoder (VAE) model with data relationship considered while generating low-dimensional latent vectors from high-dimensional raw sensory data;
    \item Automatically learn symbolic operators to identify implicit symbolic states and predict state transitions for task and motion planning. This exempts manual pre-programming and makes it possible for a robot to carry out TAMP in dynamic and uncertain environments. 
    \item To the best of our knowledge, the proposed method is the first ever autonomous battery disassembly solution that can handle dynamics and uncertainties. 
\end{enumerate}

\section{Related Work}
\noindent Disassembly strategy research ~\cite{harper2019recycling, rastegarpanah2021semi} focuses on generating a strategy to disassemble components of the battery step by step. Besides, the major obstacle towards autonomous disassembly is the inability of robots to autonomously plan in an unstructured and dynamic environment. Task and Motion Planning (TAMP) focuses on solving robotic planning problem in unstructured and dynamic environment~\cite{bonet2001planning, gizzi2019creative, sarathy2020spotter, garrett2020pddlstream}. TAMP provides a powerful means to figure out the plan for real-world task, which are both long-horizon and multi-step. The limitation of TAMP is that it relies on hand-specified planning models, such as operators in Planning Domain Definition Language (PDDL). We aim to automatically learn the planning mechanism so that a robot can adapt to environmental dynamics and uncertainties.

NeuroSymbolic~\cite{garcez2020neurosymbolic} tries to integrate high-level (symbolic) inference with lower-level (neural network) perception seamlessly to leverage the benefit of both methods. It opens a door for TAMP. Some work attempted to use networks to predict the current symbolic states~\cite{huang2019continuous, kase2020transferable, migimatsu2021grounding, ren2021neurosymbolic}. However, these work still rely on handcrafted predicates. Other approaches attempted to learn symbolic operators based on neural networks~\cite{konidaris2018skills, asai2018classical}. These work learns not only the symbolic representation from raw sensory data but also a way to ``execute'' an action in the space of the symbolic representation. The work most related to ours is the Latplan~\cite{asai2018classical}, which learns the symbolic representation in unsupervised manner and plans with the latent representation. However, all these previous work assume working on tasks with discrete and a definite number of states, which is not applicable in the battery disassembly usage scenario.

\section{Problem Definition}
\noindent This paper focuses on a typical subtask of battery disassembly: bolt removing. Disassembling bolts is straightforward for humans while challenging for robots to achieve the autonomous disassembly under dynamic and unstructured environments. The robot needs to choose sequences of actions autonomously based on current situation. 

We formulate the bolt disassembly task as a planning problem: ($S_0$,$S_G$,$A$), where $S_0$ is the initial state of the disassembly task, $S_G$ is the goal state, $A$ is the sequence of disassembly action primitives, which will be defined below. The planning algorithm aims to find a sequence of primitives $A=\{a_1,a_2,...,a_n$\}. By executing these primitives in turn, the system moves from its initial state to the goal state. 

The action primitives are defined through analyzing the process of the bolt disassembly~\cite{8954893}. There are five action primitives for the bolt disassembly task: 

\noindent\textbf{Approach}: This primitive directs the nut runner to approach the bolt, whose rough location is required to be known.

\noindent\textbf{Mate}: This primitive adjusts the location of the nut runner to get its center-line aligned with the center-line of the bolt.

\noindent\textbf{Push}: If there is no sufficient space for the nut runner to conduct disassembly, this primitive will push away obstacles around the target.  

\noindent\textbf{Insert}: This primitive directs the nut runner to rotate around the bolt axis to ensure valid contact, normally with a contact depth of $1$ mm to the gasket under the bolt.

\noindent\textbf{Disassemble}: This primitive directs the nut runner to rotate counterclockwise and move away from the bolt at a speed equal to the amount of pitch that the bolt moves per second.

\section{Proposed Method}
\begin{figure*}[ht] 
\centering 
\includegraphics[width=16cm, height=9.2cm]{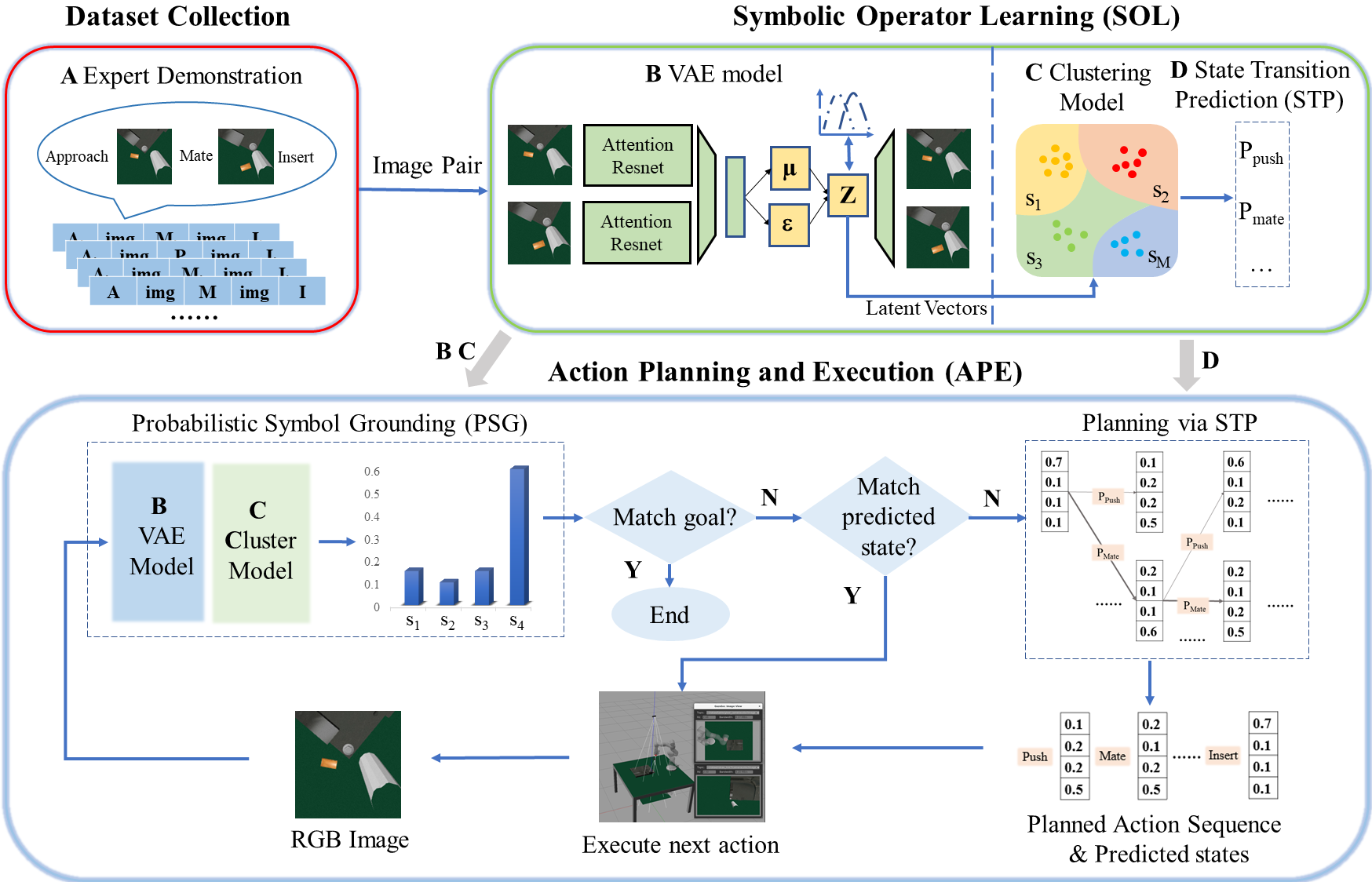}
\caption{System Overview. The dataset, shown in the Dataset Collection (red box), is a collection of action sequences from expert demonstrations and contains both action primitives and the sensory images between the action primitives. In symbolic operator learning (green box), we first train the augmented VAE network to get the latent vector for each image and then cluster them to generate the learned symbolic states. In the Action Planning and Execution (blue box), PSG maps the current image to the probabilistic symbol state distribution based on B and C. We choose to use distributions instead of a discrete symbolic state to avoid invalid results induced by thresholding. The planner searches for the best action sequence and corresponding predicted state distributions that lead to the goal in state-space via STP (State Transition Prediction). Note that the planning is online in that it will re-plan if the current state is not consistent with the original plan during execution.} 
\label{fig:method}
\vspace{-0.1cm}
\end{figure*}

\noindent The challenges of long-horizon tasks such as bolt removing involve identifying the states of the task and transitions across states with different actions taken. This paper proposes a method to map the raw sensory inputs into the symbolic space. The states and transitions between states caused by actions in the symbolic space are also defined. This way, the algorithm can do symbolic planning to reach the desired goal state. We define symbolic operators for each action primitive as: $\it o=(name, action\,primitive, parameter, $PSG$\it,$ STP$\it)$, where $\it name$ denotes the name of the operator, $\it action\,primitive$ denotes the corresponding action primitive in the physical space, $\it parameter$ describes the required sensory data of the operator, PSG denotes the probabilistic symbol grounding model that identifies the system's current state from $\it parameter$. STP denotes the state transition prediction model of the operator given the designated $\it action\,primitive$. Note that different operators have different STPs, but they may share the same PSG to map the same sensory data to a probabilistic symbolic state distribution. 

With the definition of the symbolic operators, we further propose to learn operators from demonstrations. As shown in ~\figref{fig:method}, the overall method consists of two main components: ``Symbolic Operator Learning (SOL)" for training and ``Action Planning and Execution (APE)" for inference. SOL trains the symbolic operators, which include a probabilistic symbol grounding (PSG) model and a state transition prediction (STP) model. The PSG model maps from high-dimensional sensory inputs (without losing generality, we focus on RGB images only in this paper) to a low-dimensional probabilistic symbolic state distribution. It includes an augmented VAE model and a clustering model, which are trained from the demonstration action sequences in the SOL component. STP model is also trained from demonstration action sequences to predict the state transition given an action primitive and its current probabilistic symbolic state distribution.  

The APE component is designed to find the best path from the current state to the goal state for the robot. First, the PSG model generates low-dimensional probabilistic symbolic state distribution from the input images. Based on STP model, the planner then searches for the best path to produce a valid action plan that includes both action primitives and the intermediary probabilistic symbolic state distributions. The robot executes actions in the plan consecutively and examines the current state distribution after each action. It will stop once the goal state is reached. If not, it will check if the current state distribution is as predicted by the plan. If so, it will carry on executing the next action of the plan. Otherwise it will invoke planning via STP model to yield a new action plan. This way, the method produces closed-loop policies in long-horizon robot manipulation tasks. In the following sections, we present detailed descriptions for each component.

\subsection{Dataset Collection}
\noindent The training data are demonstrations of battery bolt disassembled by an expert controlled robot. During the robotic operating process, the actions and the sensory data between actions are logged to form the action sequences. ~\tabref{tab:plain} shows a few demo action sequences. Although this paper mainly focuses on processing RGB image inputs, the dataset can accommodate heterogeneous sensory inputs, such as symbolic ones ($s_0$, $s_1$, $\ldots$). Note that for each action primitive the action sequence contains only one image,  which is captured after the leading action primitive is carried out. 

\begin{table*}[h]
\centering 
\resizebox{2.0\columnwidth}{!}{
\begin{tabular}{cccccccccccccc}
\hline

\begin{tabular}{@{}c@{}}Act \\ Seq. $\#$\end{tabular} & 
\begin{tabular}{@{}c@{}}Act \\ Seq. Type\end{tabular} &
\begin{tabular}{@{}c@{}}Sensory \\ data\end{tabular}   & 
Act & 
\begin{tabular}{@{}c@{}}Sensory \\ data\end{tabular}  &
Act & 
\begin{tabular}{@{}c@{}}Sensory \\ data\end{tabular}  &
Act & 
\begin{tabular}{@{}c@{}}Sensory \\ data\end{tabular} &
Act &
\begin{tabular}{@{}c@{}}Sensory \\ data\end{tabular} &
Act &
\begin{tabular}{@{}c@{}}Sensory \\ data\end{tabular}\\
\hline
1 & AID     &$s_0$ &Approach & $img_1$ & Insert & $s_1$ & Disassemble & $s_2$   \\
2 & AMID    &$s_0$ &Approach & $img_2$ & Mate & $img_3$ & Insert   & $s_1$ & Disassemble & $s_2$   \\
3 & AID     &$s_0$ &Approach & $img_4$ & Insert & $s_1$ & Disassemble & $s_2$   \\
4 & APID    &$s_0$ &Approach & $img_5$ & Push & $img_6$ & Insert  & $s_1$ & Disassemble & $s_2$   \\
5 & AMID    &$s_0$ &Approach & $img_7$ & Mate & $img_8$ & Insert  & $s_1$ & Disassemble & $s_2$    \\
6 & APMID   &$s_0$ &Approach & $img_9$ & Push & $img_{10}$ & Mate & $img_{11}$ & Insert    & $s_1$ & Disassemble & $s_2$  \\
\hline
\multicolumn{6}{l}{$s_0$ : get a coarse position of bolt}& 
\multicolumn{6}{l}{$s_1$ : the screw(bolt) head is fitted over by a wrench socket}\\
\multicolumn{6}{l}{$s_2$ : the screw(bolt) is removed}&
\multicolumn{6}{l}{$img_{i}$ : the image captured by an ``eye-in-hand" camera}\\
\hline
\end{tabular}}
\caption{Action Sequences Obtained from Demonstrations.}
\label{tab:plain}
\vspace{-0.2cm}
\end{table*}

\subsection{Symbolic Operator Learning}
\noindent This model learns the symbolic operators described above. The objective of symbolic operator learning is to first learn a latent vector for each raw input using the augmented VAE model. The symbolic states are then obtained by the clustering model. These two learned models later will be used to generate the probabilistic symbolic state distribution to enable the task and motion planning. 
\subsubsection{Augmented Variational Autoencoder}

\noindent Given the training dataset, we need to learn to capture low-dimensional state representations from the high-dimensional RGB image data. Later these representations are clustered by the clustering model to learn salient symbolic states to evaluate state transitions and the goal state. Regular variational autoencoder VAE~\cite{kingma2013auto} is widely used to learn a mapping $q(z \vert x)$ from high-dimensional input {\it x} to a low-dimensional latent vector {\it z}. However, it only considers individual images, while relationships between images are critical in our task. Specifically, in our framework it is desirable to have the distance between the latent vectors of two input images as small as possible if they trigger the same action, or as large as possible if they trigger different actions. Only in this way, the system can cluster all related images into similar states and identify necessary actions based on these states. Therefore, we design and  train an augmented VAE to address the symbolic state learning. 

\noindent \textbf{Network Structure:} A VAE consists of both encoder and decoder neural networks. We apply a VAE with a Resnet architecture~\cite{he2016deep}. When training the encoder neural network, CBAM~\cite{woo2018cbam}, a form of self-attention, is added to emphasize meaningful features along channel and spatial axes of input images. A residual attention network is applied as the encoder network, while a symmetric network without the attention module is used as decoder network.

\noindent \textbf{Training Loss:} The loss function of regular VAE is defined as:
\begin{equation}{
\begin{aligned}
    \mathcal{L}_{\text {vae}}(x)=E_{z \sim q(z \vert x)}[\log p(x \vert z)]+\beta \cdot D_{K L}(q(z \vert x) \| p(z))
\end{aligned}
\label{equ:1}}
\end{equation}

\noindent where the first term formulates the reconstruction loss between the reconstructed output $\hat x$ and the given input {\it x}. The second term, also called KL divergence term, regularises the returned distribution $q(z \vert x)$ to ensure a better organisation of the latent space. The hyper-parameter $\beta$ is adjusted to achieve a more or less regularized latent space according to its value as in~\cite{Higgins2017betaVAELB}.

However, this loss function only leverages the information from individual raw pixels. To take full advantage of the information in demonstrations, we extend the loss function to include relationship between inputs as follows:
\begin{equation}
\begin{aligned}
    \mathcal{L}\left(\mathrm{I_{1}}, \mathrm{I_{2}}\right)=\frac{1}{2}\left(\mathcal{L}_{\text {vae}}\left(\mathrm{I_{1}}\right)+\mathcal{L}_{\text {vae}}\left(\mathrm{I_{2}}\right)\right)+\alpha \mathcal{L}_{\text {r}}\left(\mathrm{I_{1}}, \mathrm{I_{2}}\right)
\end{aligned}
\label{equ:2}
\vspace{-0.05cm}
\end{equation}

As the major extension, $L_r$ is added to quantify expectations of the task planning system to consider the relationship between images. $L_r$ is defined in \equref{equ:3}. The $\alpha$ could be tuned to adjust the weight of distance loss compared to regular VAE encoding-decoding loss.
\begin{equation}
\begin{split}
\mathcal{L}_{\text{r}}&\left(\mathrm{I}_{1}, \mathrm{I}_{2}\right)= \\
&\begin{aligned}
    \begin{cases}\max \left(0,2 * d_{m}-\left\|z_{1}-z_{2}\right\|_{1}\right)  &\text {if exclusive} \left( \mathrm{I}_{1}, \mathrm{I}_{2}\right)\\ \max \left(0, d_{m}-\left\|z_{1}-z_{2}\right\|_{1}\right) 
    & \text {if independent} \left( \mathrm{I}_{1}, \mathrm{I}_{2}\right) \\ \left\|z_{1}-z_{2}\right\|_{1}
    & \text {if inclusive} \left( \mathrm{I}_{1}, \mathrm{I}_{2}\right)\end{cases}
\end{aligned}
\end{split}
\label{equ:3}
\end{equation}

In \equref{equ:3}, $z_1$ and $z_2$ are the latent vectors encoded from input images $I_1$, $I_2$, respectively. The hyper-parameter $d_m$ describes the extended distance margin of the latent vectors, which helps enforce $z_1$ and $z_2$ distribute in separate regions in the latent space. As \equref{equ:3} shows, we divide the relationship between pairs of images in demonstrations into three categories: inclusive, exclusive and independent.

The first category, $inclusive(img_i, img_j)$, means that two images share identical subsequent action sequences in demonstrations. In fact, the identical subsequent action sequence indicates that the two images reflect the same state, so it is desirable to make their latent vectors as close as possible. Based on this relationship, we can get $set_{inclusive}$ for each image, which contains all the images that have the same subsequent action sequences in training. The second category, $exclusive(img_i, img_j)$ \footnote{In our definition, $exclusive$ relationship has a higher priority over $inclusive$. In other words, if we find that two images are both $inclusive$ and $exclusive$, the loss value is calculated according to the $exclusive$ relationship.}, means that two images appear at different locations in the same action sequence. The $set_{exclusive}(img_i)$ also includes all images that are exclusive with any image in $set_{inclusive}(img_i)$. In fact, two exclusive images indicate that the latent vectors of these two images should be distinguishable, based on which the planner can make a correct decision. Therefore, it is desirable to make the distance between the latent vectors of two exclusive images as far as possible. If the relationship falls to neither of the two categories, it belongs to the third category, $independent(img_i, img_j)$. When performing VAE encoding, we expect the latent vector distances of $independent$ images to be between the distances of $inclusive$ and $exclusive$.

\subsubsection{Clustering Model} 

\noindent The clustering model takes the latent vectors encoded by the augmented VAE to learn symbolic states. It is based on the intuition that the latent vectors should distribute in disjoint clusters if they belong to different symbolic states. We apply the unsupervised K-means algorithm to partition unlabelled data into a certain number of distinct groupings in the latent space where the symbolic state regions are identified. The primary challenge here is to determine the optimal number of clusters. A 3-step algorithm is designed to determine the best number of symbolic states, which is the number of clusters in the learning process.

\noindent 1. For all training data, apply the k-means clustering to the latent vectors with varying $k$ to obtain the cluster labels, which are also the labels of the symbolic states the corresponding images belong to. 

\noindent 2. If different images in the same action sequence end up being mapped to the same symbolic states, then the classification of the symbolic states in this sequence is considered to be ``incorrect” since their symbolic states in the same sequence should all be different. We then calculate the number of incorrect cases for each model using different $k$ values.

\begin{figure}[t]
\centering 
\subfigure[Incorrect sequences versus k]{
\label{Fig.sub.1}
\includegraphics[width=4.2cm,height = 2.8cm]{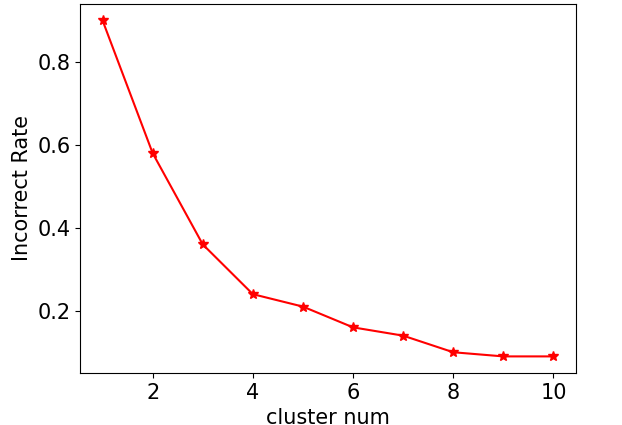}}\subfigure[Success rate versus k]{
\label{Fig.sub.2}
\includegraphics[width=4.2cm,height = 2.8cm]{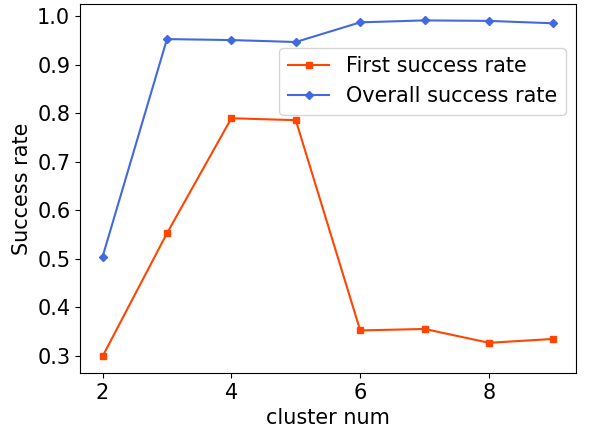}}
\vspace{-1\baselineskip}
\caption{(a) The number of incorrect sequences decreases as $k$ increases, and stops to decrease significantly at $4$. (b) The first and overall success rates both reach relatively good performance with $k$ values of $4$ or $5$.}
\label{1}
\vspace{-0.15cm}
\end{figure}

\noindent 3. Plot a curve of the number of incorrect cases versus the value of $k$ as shown in~\figref{Fig.sub.1}. The value of $k$ with the error rate of the curve dropping below a predefined threshold is then chosen. This criterion implies that increasing $k$ beyond the chosen value only brings trivial performance gain, hence should be avoided.

Through experiments, we find that $4$ is the appropriate number of symbolic states in this task. Then, we assume each cluster is produced by a multivariate normal distribution. The clusters can be modeled as a mixture of different Gaussian distributions based on which we can calculate the probabilistic symbolic state distribution for each new incoming image. The planning via STP module takes in the probabilistic distribution to search the optimal execution plan to reach the goal state. The plan includes both the action primitives and the intermediary probabilistic symbolic state distribution at each step, as shown in ~\figref{fig:method}.

\subsection{Action Planning and Execution}
\noindent The action planning and execution component first maps all incoming raw images to a probabilistic symbolic state distribution by PSG. The action plans are then generated based on the trained STPs. 

\subsubsection{Probabilistic Symbol Grounding}
\noindent PSG is designed to map an image to a low-dimensional probabilistic symbolic state distribution (shown as the bar chart in ~\figref{fig:method}). Each incoming image is first mapped to the latent space. Then the clustering model generates a discrete probabilistic distribution across all possible symbolic states using a Gaussian Mixture Model (GMM). Thus for image at time $t$, we can get the corresponding symbolic state distribution $S_t=[p_{s_1}, p_{s_2}, ..., p_{s_M}]^T$, assuming total $M$ symbolic states.

\subsubsection{Planning via State Transition Prediction}
\noindent With the probabilistic symbolic state distribution as its input, a planner searches for the best action sequence to reach the goal state. A traditional way is to use the PDDL-based planner, whose basic idea is to explore the currently applicable action sequence and end when the predicted state matches the goal state. However, traditional PDDL-based planner requires a deterministic state as input. Therefore, the probabilistic symbolic state distribution defined in the previous section can not be used directly. To this end, we propose a probabilistic forward state-space search planner. Details of the state transition and goal satisfaction are described in following sections. 

\noindent \textbf{State Transition:} It is necessary to model the state transition to conduct probabilistic planning. We define images at the same position in identical action sequence as a group. For example, $img_1$ and $img_4$ in ~\tabref{tab:plain} are in the same group because they are in the identical action sequence ($AID$) and at the same position (between $Approach$ and $Insert$). We assume that there are $N$ groups in the training set.  Based on the definition of the group, we define a $M \times N$ matrix $N_{cs}$, where $M$ is the number of symbolic states, $N$ is the number of groups. The entry $N_{cs}^{ij}$ is the number of images of the $i$-th symbolic state in the $j$-th group in the training set. Then we normalize the $N_{cs}$ matrix along the row and column dimensions respectively to get matrix $Q$ and $K$. $Q_{ij}$, the entry of $Q$, represents the $j$-th group percentage (purity) in the $i$-th symbolic state. $K_{ij}$, the entry of $K$, represents the $j$-th symbolic state percentage (purity) in the $i$-th group. We also define a $N \times N$ transition matrix $T$ for each action whose entry is a binary value representing the feasibility of the transition between the two groups by the action. For example, we have $T^{Push}$ for action $Push$ and $T^{Mate}$ for action $Mate$, etc.

The primary challenge is to predict the categorical symbolic state distribution for each action given the current symbolic state distribution. Given the definition of the matrices, we can now calculate the probability of moving from the $i$-th symbolic state to the $j$-th symbolic state with any action:
\begin{equation}
\begin{aligned}
    P_{ij}^{a}= \sum_{m=1}^{N} \sum_{n=1}^{N} Q_{i m} \cdot T_{m n}^{a} \cdot K_{j n}
\end{aligned}
\label{equa:single_transit}
\end{equation}
where $a$ stands for the type of action. Then we can build a matrix $P^{a}$ where the entry is calculated by Equ.(\ref{equa:single_transit}). The successor state distribution is then predicted as:
\begin{equation}
\begin{aligned}
    S_{t+1}= S_{t} \cdot P^{a}
\end{aligned}
\label{equa:transit}
\end{equation}
where $S_{t}$ and $S_{t+1}$ are the current symbolic state distribution and the predicted state distribution, respectively, with regards to action $a$.

It should be pointed out that the process of building a matrix $P^{a}$ applies not only to image data but also to sensory data that can be directly symbolized, such as $s_0, s_1$ in ~\tabref{tab:plain}. STP model can be easily generated by extending $Q$, $K$, and $T$ to accommodate different kinds of sensory inputs.

\noindent \textbf{Goal State Satisfaction:} For each planning step, it is necessary to evaluate if the predicted state distribution matches the goal distribution. We address this by quantitatively measuring the KL-divergence between the two distributions. The lower the KL-divergence, the more possible for the plan to reach the goal state. Thus, a plan is deemed applicable if the KL-divergence drops below a predefined threshold $\varepsilon$.

\section{Experiments}
\begin{figure} 
\centering \centering 
\includegraphics[width=1.0\columnwidth]{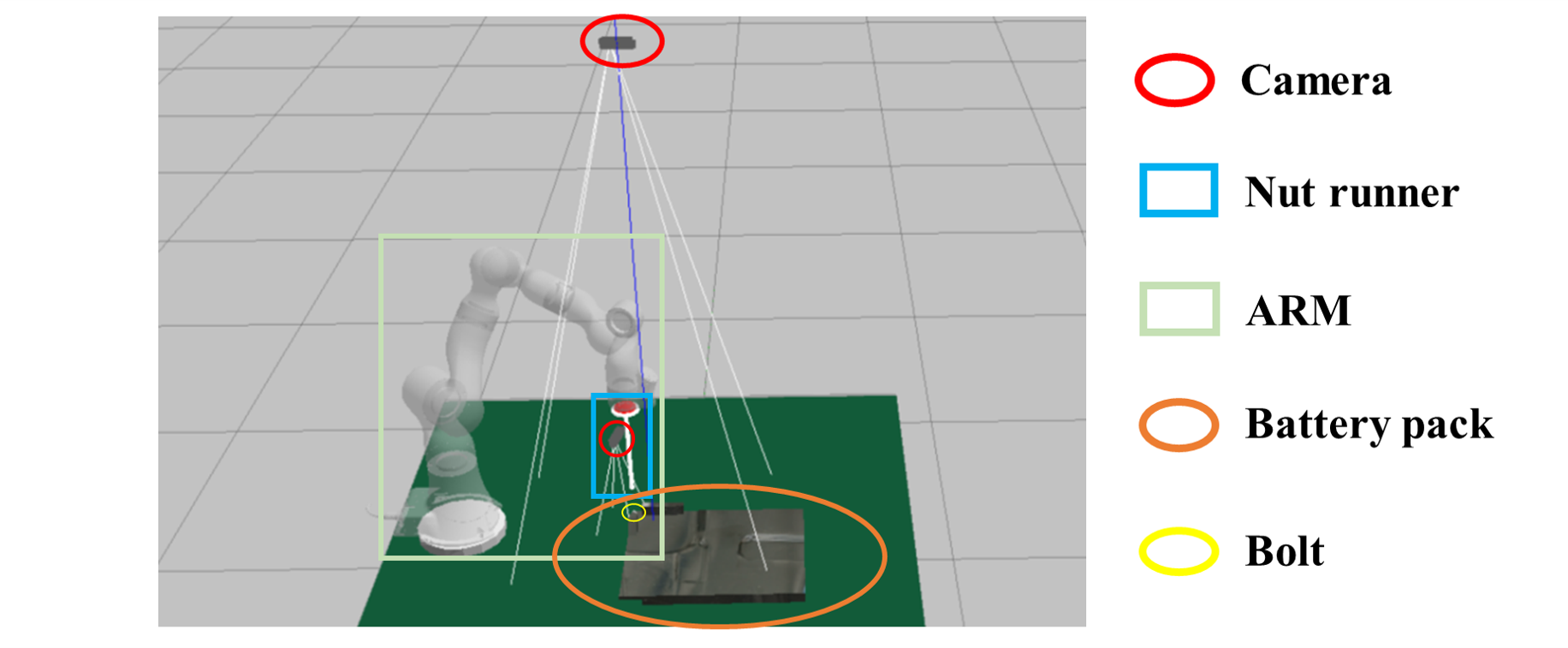}
\vspace{-0.5cm}
\caption{The Disassembly Platform}\label{fig:platform}
\vspace{-0.1cm}
\end{figure}

\noindent In this section, we evaluate the performance of the proposed method with the battery bolt disassembly task. \figref{fig:platform} shows the test environment. We use the simulation environment Gazebo with a 7-degrees-of-freedom robot arm as the testing platform. The robot arm is equipped with a nut runner at its end effector. In addition, there is a global camera with a top-down view and an eye-in-hand camera. Intel RealSense Depth Camera D435i is used in experiments. The simulator is also configured with a Chang’an motors EADO EV460 battery pack, bolts, and a woodblock as the obstacle. The experiments are designed to answer two questions: (1) How well does the proposed method accomplish the battery bolt disassembly task? (2) How effective does the method handle dynamics and uncertainties? To simulate the dynamically changing environments, obstacles are randomly added around the bolts to verify if the robot can adjust under different circumstances.

\subsection{Static Environment}
\noindent In this experiments, no obstacle exists in the environment. The robot makes decisions autonomously in multiple sets of environment to accomplish the disassembly tasks. Our simulation implementation focuses on image related steps and assumes the success of the task once the nut runner fits over the bolt. We assume that there is no other external environment changing force other than the robot itself.~\tabref{tab:sexperiment} shows the results of the experiments. The first success means the execution succeeds with planning only once. The rectified success means that although the first plan deviates from the best path, the planner successfully adjusts the plan online during execution and finally reaches the goal state. The experiments show that the proposed method is highly effective with a success rate (SR) of $79.3\%$ and an  overall success rate of $95.4\%$. 

\begin{table}[h]
\centering 
\resizebox{1.0\columnwidth}{!}{
\begin{tabular}{ccccc}
\hline
   & First SR(\%) & Rectified SR(\%) & Overall SR(\%) & Num\\
\hline
AI     & 89.4 & 10.5 & 99.9 & 1002    \\
API    & 80.9 & 13.9 & 94.8 & 523     \\
AMI    & 74.2 & 17.1 & 91.3 & 952     \\
APMI   & 67.6 & 27.5 & 95.1 & 527     \\
\hline
summary   & 79.3 & 16.1 & 95.4 & 3004     \\
\hline
\end{tabular}} 
\caption{Result of task planning in a static environment}
\label{tab:sexperiment}
\end{table}

We also verify the influence of the number of clusters on the final success rates in this experiment. The results are shown in the \figref{Fig.sub.2}. It can be seen a  large cluster number leads to a significant drop of the first success rate, although the overall success rate does not change significantly. The reason is that when the number of clusters exceeds the intrinsic number of categories in the system, the original images in the same category will be divided into different clusters, which leads to wrong decisions of the subsequent actions and deterioration of the first success rate.

\subsection{Dynamic Environment}
\begin{figure}[t]
\centering \centering \vspace{-0.40cm}
\subfigure[]{
\label{Fig.test.sub.1}
\includegraphics[width=4.3cm,height = 2.8cm]{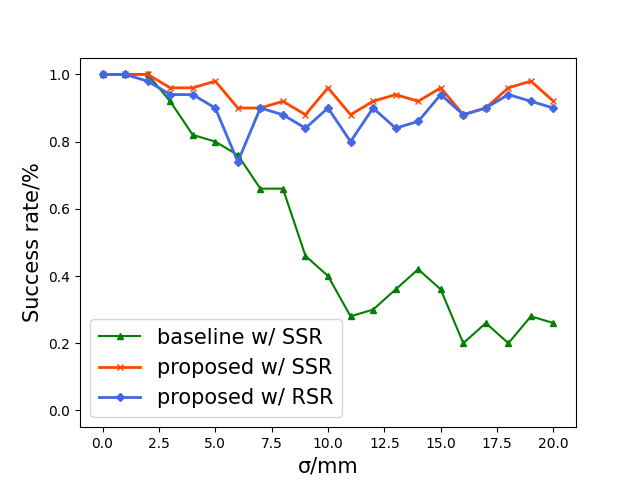}}\subfigure[]{
\label{Fig.test.sub.2}
\includegraphics[width=4.3cm,height = 2.8cm]{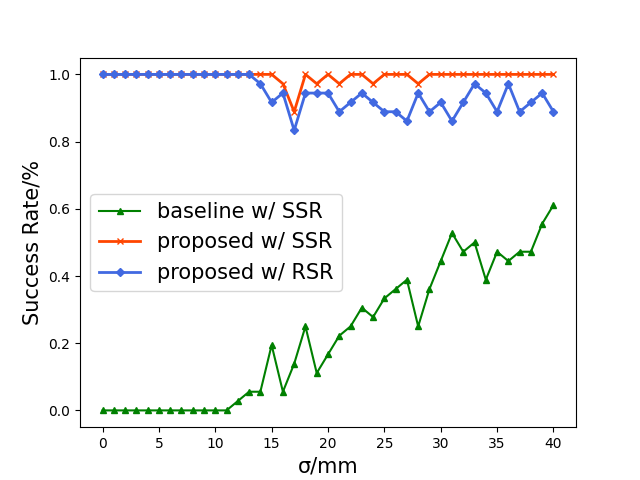}}
\vspace{-0.45cm}
\caption{Result of task planning in dynamic environment. (a) success rate of random bolt position tests; (b) success rate of random obstacle tests.}
\label{fig:result}
\vspace{-0.2cm}
\end{figure}

\noindent Different from previous experiments, we introduce dynamics and uncertainties in the dynamic environments. Specifically, we present two typical noises; one is that the position of the bolt is shifted by a random distance of a $N(0,\sigma^2)$ distribution, and the other is that there is an obstacle near the bolt (at the position with a $N(0,\sigma^2)$ distribution to the bolt position). These two situations cover two typical dynamic problems in the bolt disassembly environment. The first problem concerns cases of inaccurate bolt positions due to an out-of-date knowledge base, less than optimal perception, flawed robot executions, etc. Our planning needs to accommodate these inaccuracies. The second case concerns incidental obstacles around the bolt due to not-cleared-up debris, fallen-off parts, or other issues. In these cases, the robot needs to determine whether the new obstacle will affect the subsequent operations according to current situations. If necessary, the method automatically introduces $Push$ operation to clear the barriers. \figref{fig:result} shows the result of these experiments. The baseline is an existing pre-programmed method that mandates accurate enough bolt positions from perception without accommodating environment variations. It suffers from online dynamics and uncertainties, which makes it rely on human intervention. Our proposed method is able to adjust according to environment dynamics and uncertainties. We define two success metrics. A Standard Success Rate (SSR) metric counts all cases as successful as long as the robot completes the task successfully. A Rigorous Success Rate (RSR) metric counts a case as successful only when no redundant actions are taken in the plan. Otherwise the task is considered to fail even if the robot does complete the task finally. On average, our approach can achieve a SSR of $93\%$ in the random bolt position test and $99\%$ in random obstacle test. As \figref{fig:result} shows, even using RSR, our system still performs effectively ($90\%$ and $94\%$ success rates on average).     

\section{Summary and Future Work}
\noindent The performance of pre-programming methods deteriorates in complex and unstructured working environments due to the high uncertainties, which makes human intervention mandatory. This paper addresses the problem by learning symbolic operators to enable autonomous TAMP.  Extensive experiments show that our method can autonomously and efficiently handle typical usage scenarios in the battery bolt disassembly tasks. In future work, we plan to implement the system on a real-world disassembly platform. We also consider enabling more operators to support upper cover removal, bus bar removal, etc. Moreover, this method can be extended to tasks beyond battery disassembly. To further improve the practicality of the system, it will be helpful to study how to optimize the two components of the proposed method: SOL and APE. For example, one-shot or few-shot learning can be considered to learn symbolic operators with fewer data.

\newpage
\bibliographystyle{named}
\bibliography{ijcai22}

\clearpage
\appendix

\section{Appendix A}
\subsection{Implementation Details}
\subsubsection{Network and Training}


We generate $2000$ action sequences from expert demonstrations. After the raw sensory inputs are obtained, the center image regions are cropped into a size of $180 \times 180$. The cropped images are then resized into $256 \times 256$ and fed into our proposed network for symbolic operator learning. We apply the augmented VAE with a Resnet architecture for the encoder and decoder networks where a $64$-dimensional latent space is obtained. The concrete structure of our network is tabulated in~\tabref{tab:network}. 

\begin{figure}[h]
\centering
\subfigure[]{
\includegraphics[width=\columnwidth]{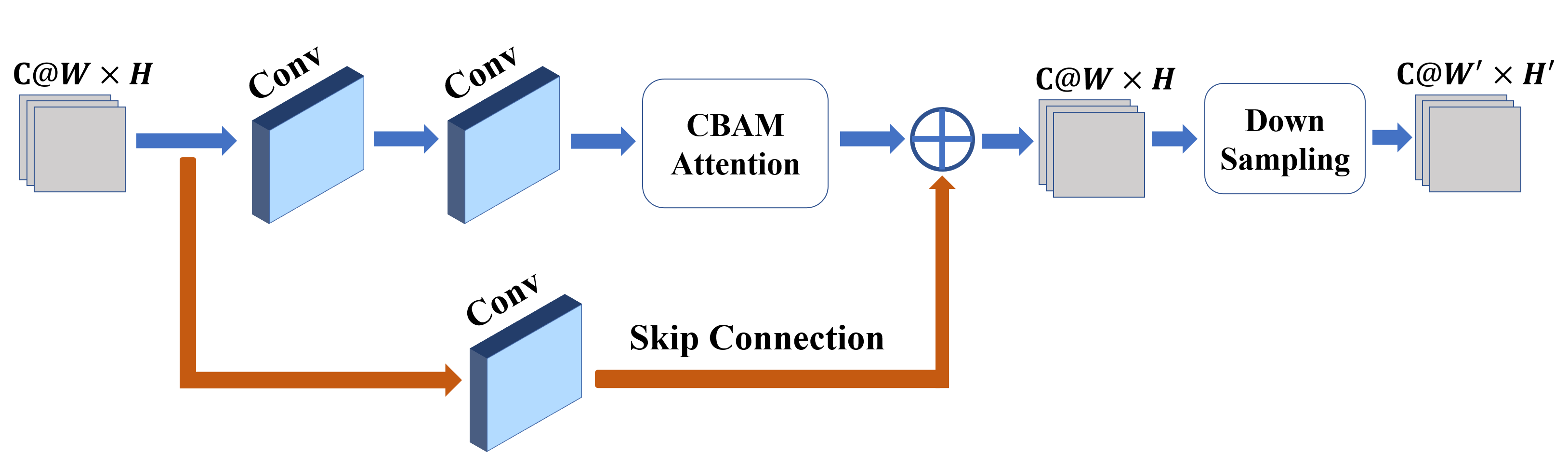}}
\subfigure[]{

\includegraphics[width=\columnwidth]{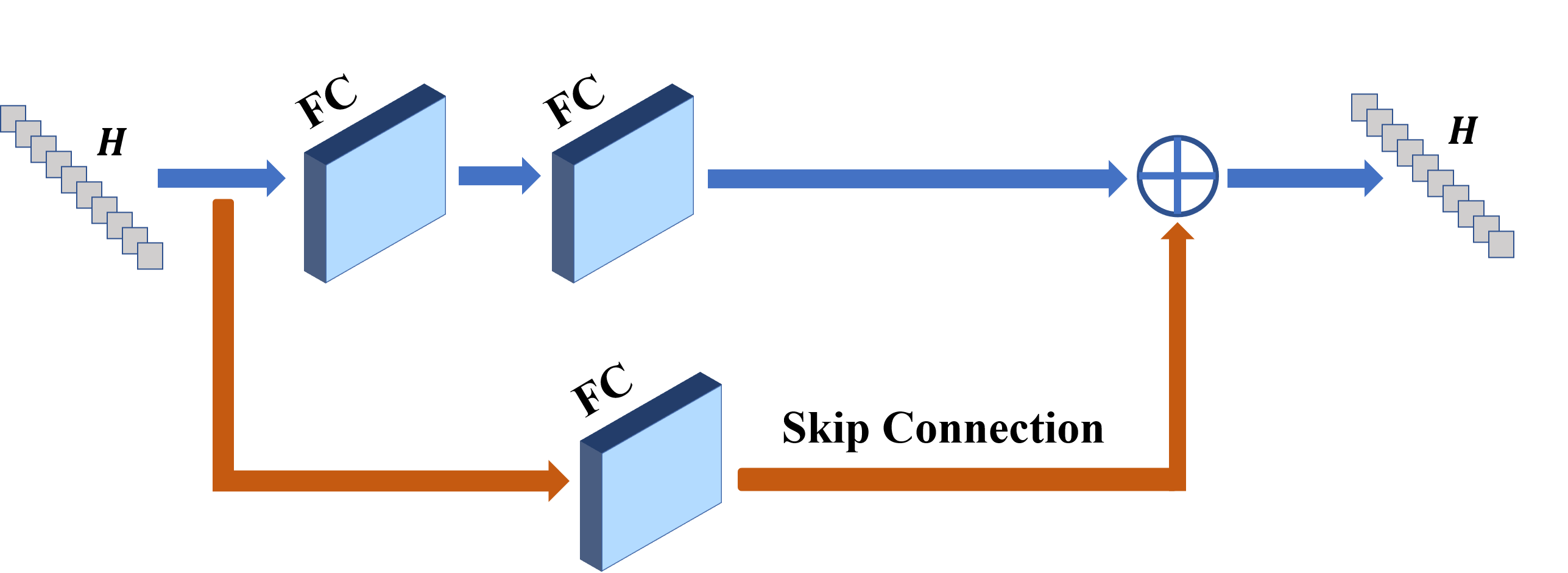}}
\caption{(a) Residual attention module, the $W^{'}$ and $H^{'}$ we use is $W/2$ and $H/2$. (b) FC residual module, the input and output share same dimension.}
\label{fig:module}
\vspace{-0.2cm}
\end{figure}

We incorporate residual attention module (RAM) and FC residual module (FRM) into our network. \figref{fig:module}(a) shows the structure of the residual attention module (RAM) used in the network. Each module comprises a residual layer which contains $2$ convolutional blocks, a CBAM attention block, and a skip connection which only contains $1$ convolutional block. The two connections are followed by a downsampling block with a stride of $2$. The downsampling block is replaced with a transposed convolutional upsampling block in the decoder network. The FC residual module is shown in \figref{fig:module} (b), in which the attention and upsampling block are removed, and convolutional blocks are replaced with fully connected layers.

\begin{table}[h]
\tiny 
\centering
\resizebox{1.0\columnwidth}{!}{
\begin{tabular}{ccc}
\hline
Layer & Layer Structure & Output Size\\
\hline
Conv header & C8, (3,1,1) & $8\times256\times256$    \\
Group \#1 & RAMs(n=6) & $256\times8\times8$     \\
Avg pool & 3 & $256\times2\times2$     \\
Flatten & None & 1024     \\
Group \#2 & FRMs(n=1) & 1024     \\
FC(mean/var) & [1024, 64]/[1024, 64] & 64/64     \\
FC & [64, 4096] & 4096     \\
Unflatten & None & $1024\times2\times2$     \\
Group \#3 & RAMs(n=7) & $8\times256\times256$     \\
Conv & C3, (3,1,1) & $3\times256\times256$     \\

\hline
\end{tabular}} 
\caption{Proposed Network Architectures. RAMs and FRMs are the convolutional and connected residual attention modules illustrated in \figref{fig:module}. $n$ is the number of modules in each group. In Conv layers, $C$ is the number of output channels, (.,.,.) means (kernel size, stride, padding). In fully connected layers, [.,.] is the input and output dimension.}
\label{tab:network}
\vspace{-0.2cm}
\end{table}

The VAE model is trained with Pytorch (version $1.9.0$) on NVIDIA Geforce RTX 3060 Ti. The loss is minimized during the training stage using the Adam optimizer with a learning rate of $0.0001$. The total number of epochs is $75$. A batch of $24$ samples is fed into the model at each iteration. We augmented the samples by adding random color jitters of brightness (with a value from $0$ to $0.04$) and contrast (with a value from $0$ to $0.08$). Once an epoch is completed, the VAE loss defined in \equref{equ:3} is obtained and back propagates.

\subsubsection{Planner and Execution}

The algorithm of our forward state space planner is shown in \AlgRef{alg:plan}. This planner follows the same searching strategy as traditional PDDL-based planner, but takes in probabilistic distributions as its input. Given the initial state and goal state, it searches the first action in a searching queue (line $7$) by predicting the state after applying the action (line $9$). Furthermore, it extends the searching queue by a breadth-first strategy among the available actions (line $13-16$), and end until the goal state is reached (line $10$). The planner returns the planned action sequence and the corresponding predicted states between each pair of actions in the action sequence (line $11$).

\begin{algorithm}[htbp]
\caption{Plan}
\label{alg:algorithm}
\textbf{Input}: $S_{init}, S_{goal}$\\
\textbf{Parameter}: $\varepsilon, actions$
\begin{algorithmic}[1] 
\STATE Initialize $action_q, plan_q, S_q$ as $Queue$
\STATE $action_q.push(None)$
\STATE $plan_q.push([])$
\STATE $S_q.push([S_{init}])$
\WHILE{$S_q \neq \emptyset$}
\STATE $S_{cur} \leftarrow pop(S_q)$
\STATE $action_{cur} \leftarrow pop(action_q)$
\STATE $plan_{cur} \leftarrow pop(plan_q)$ \\ {\footnotesize $\triangleright$ State Prediction}
\STATE $S_{predict} \leftarrow predict(S_{cur}[-1], action_{cur})$ \\ {\footnotesize $\triangleright$ return plan and predicted states}
\IF{$KL(S_{predict}, S_{goal}) < \varepsilon$}
\STATE \textbf{return} $plan_{cur}, S_{cur}$
\ENDIF
\FOR{\textbf{each} $action$ $\in$ $actions$}
\STATE $action_q.push(action)$
\STATE $plan_q.push(Concat[plan_{cur}, action])$
\STATE $S_q.push(Concat[S_{cur}, S_{predict}])$
\ENDFOR
\ENDWHILE

\end{algorithmic}
\label{alg:plan}
\end{algorithm}

The execution algorithm is shown in \AlgRef{alg:system}. The algorithm first plans using \AlgRef{alg:plan} with the initial and goal state generated by the learned operators (line $1-2$) to produce an action plan and predicted states (line $4$). Then, the actions in the action plan are executed (line $8$) one by one. During the execution, the actual state is checked (line $8-11$). If the actual state is found to be different from the corresponding predicted state (line $11$), the system re-plans with the current state (line $12$). The system keeps following this process until the goal state is reached (line $5$).

\begin{algorithm}[htbp]
\caption{Execution}
\label{alg:algorithm}
\textbf{Input}: $img_{init}, S_{goal}$\\
\textbf{Parameter}: $\varepsilon$
\begin{algorithmic}[1] 
\STATE $z_{init} \leftarrow Encoder(img_{init})$ {\footnotesize $\triangleright$ encode img to latent vector}
\STATE $S_{init} \leftarrow Prob\_SG(z_{init})$ {\footnotesize $\triangleright$ probabilistic symbol grounding}
\STATE $kl := KL(S_{init}, S_{goal})$
\STATE $plan, S_{predict} := plan(S_{init}, S_{goal})$ {\footnotesize $\triangleright$ apply Algorithm 1}
\WHILE{$kl > \varepsilon$}
\STATE $action \leftarrow pop(plan)$  
\STATE $S_{pre\_cur} = pop(S_{predict})$ \\ \footnotesize $\triangleright$ apply action and return successor image
\STATE $img_{cur} \leftarrow Execute(action)$ 
\STATE $z_{cur} \leftarrow Encoder(img_{cur})$
\STATE $S_{cur} \leftarrow Prob\_SG(z_{cur})$

\IF {$KL(S_{cur}, S_{pre\_cur}) > \varepsilon$}
\STATE $plan, S_{predict} := plan(S_{cur}, S_{goal})$ 
\ENDIF

\STATE $kl := KL(S_{cur}, S_{goal})$
\ENDWHILE
\end{algorithmic}
\label{alg:system}
\end{algorithm}

\end{document}